\documentclass[runningheads]{llncs}
\usepackage{color}
\usepackage{graphicx,wrapfig,lipsum}
\usepackage{amsmath}
\usepackage{graphicx}
\usepackage{caption}
\usepackage{enumitem}   
\usepackage{amssymb}
\usepackage{hyperref}
\usepackage{wrapfig}
\usepackage{array, caption, floatrow, tabularx, makecell, booktabs}%
\setcellgapes{3pt}
\usepackage [english]{babel}
\usepackage [autostyle, english = american]{csquotes}
\MakeOuterQuote{"}
\usepackage{csquotes}
\usepackage[boxed, ruled, vlined, linesnumbered,titlenumbered]{algorithm2e}
\renewcommand{\L}[0]{\mathcal{L}}
\newcommand{\A}[0]{\mathcal{A}}
\newcommand{\K}[0]{\mathcal{K}}

\newcommand{\Diff}[0]{{\rm Diff}}

\newcommand{\Reg}[0]{{\rm Reg}}

\begin{document}
\title{Bayesian Atlas Building with Hierarchical Priors for Subject-specific Regularization}

\titlerunning{Bayesian Atlas Building with Subject-specific Regularization}

\author{Jian Wang \inst{1}\and
Miaomiao Zhang\inst{1, 2}}

\authorrunning{Wang et al.}

\institute{ Computer Science, University of Virginia, USA \and
Electrical and Computer Engineering, University of Virginia,USA}

\maketitle   
\begin{abstract}
This paper presents a novel hierarchical Bayesian model for unbiased atlas building with subject-specific regularizations of image registration. We develop an atlas construction process that automatically selects parameters to control the smoothness of diffeomorphic transformation according to individual image data. To achieve this, we introduce a hierarchical prior distribution on regularization parameters that allows multiple penalties on images with various degrees of geometric transformations. We then treat the regularization parameters as latent variables and integrate them out from the model by using the Monte Carlo Expectation Maximization (MCEM) algorithm. Another advantage of our algorithm is that it eliminates the need for manual parameter tuning, which can be tedious and infeasible. We demonstrate the effectiveness of our model on 3D brain MR images. Experimental results show that our model provides a sharper atlas compared to the current atlas building algorithms with single-penalty regularizations. Our code is publicly available at \url{ https://github.com/jw4hv/HierarchicalBayesianAtlasBuild}.
\end{abstract}

\section{Introduction}
Deformable atlas building is to create a “mean” or averaged image and register all subjects to a common space. The resulting atlas and group transformations are powerful tools for statistical shape analysis of images~\cite{hong2017fast,ma2008bayesian}, template-based segmentation~\cite{pohl2006bayesian,rohlfing2004evaluation,iglesias2012incorporating}, or object tracking~\cite{lorenzo2002atlas,liao2019temporal}, just to name a few. A good quality of altas heavily relies on the registration process, which is typically formulated as a regularized optimization to solve~\cite{ashburner2011diffeomorphic,beg2005computing,vialard2012,zhang2019fast}. An issue in the current process of registration-based atlas construction is how to regularize model parameters. Having an appropriate regularization is critical to the "sharpness" of the atlas, as well as ensuring a set of desirable properties of transformations, i.e., a smooth and invertible smooth mapping between images, also known as diffeomorphisms, to preserve the topology of original images.

Current atlas building models either exhaustively search for an optimal regularization in the parameter space, or treat it as unknown variables to estimate from Bayesian models. While ad hoc parameter-tuning may yield satisfactory results, it requires expert domain knowledge to guide the tuning process~\cite{joshi2004unbiased,vialard2011diffeomorphic,ma2008bayesian,wang2019data}. Inspired by probabilistic models, several works have proposed Bayesian models of atlas building with automatically estimated regularizations ~\cite{allassonniere2007towards,allassonniere2008stochastic,zhang2013bayesian}. These approaches define a posterior distribution that consists of an image matching term between a deformed atlas and each individual as a likelihood, and a regularization as a prior to support the smoothness of transformation fields. The regularization parameter is then jointly estimated with atlas after carefully integrating out the image deformations using Monte Carlo sampling. However, sampling in a high-dimensional transformation space (i.e., on a dense 3D image grid $128^3$) is computationally expensive and often leads to a long execution time with high memory consumption. More importantly, the aforementioned methods are limited to regularizations with single-penalty for population studies. This prohibits the model's ability to adaptively search for the best regularization parameter associated with an individual subject, which is critical to images with various degrees of geometric transformations. The typical “one-fits-all” fails in cases where large geometric variations occur, i.e., brain shape changes of Alzheimer's disease group. Allowing the subject-specific (data-driven) regularization can substantially affect the sharpness and quality of the atlas~\cite{yeo2008effects}. 
 
In this paper, we propose a hierarchical Bayesian model of atlas building with subject-specific regularizations in the context of Large Deformation Diffeomorphic Metric Mapping (LDDMM) algorithm~\cite{beg2005computing}. In contrast to previous approaches treating the regularization of individual subjects as a single-penalty function with adhoc parameters, we develop a data-adaptive algorithm to automatically adjust the model parameters accordingly. To achieve this, we introduce a novel hierarchical prior that features (i) prior distributions with multiple regularization parameters on the group transformations in a low-dimensional bandlimited space; and (ii) a hyperprior to model the regularization parameters as latent variables. We then develop a Monte Carlo Expectation Maximization (MCEM) algorithm, where the expectation step integrates over the regularization parameters using Hamiltonian Monte Carlo (HMC) sampling. The joint estimation of model parameters including atlas, registration, and hyperparameters in the maximization step successfully eliminates a massive burden of multi-parameters tuning. We demonstrate the effectiveness of our algorithm on both 2D synthetic images and 3D real brain MRIs. 

To the best of our knowledge, we are the first to extend the atlas building to a data-adaptive and parameter-tuning-free framework via hierarchical Bayesian learning. Experimental results show that our model provides an efficient atlas construction of population images, particularly with large variations of geometric transformations. This paves a way for an improved quality of clinical studies where atlas building is required, for example, statistical shape analysis of brain changes for neurodegenerative disease diagnosis~\cite{hong2017fast}, or atlas-based segmentation for in-utero placental disease monitoring~\cite{liao2019temporal}.   

\section{Background: Atlas Building with Fast LDDMM}
We first briefly review an unbiased atlas building algorithm~\cite{joshi2004unbiased} based on Fourier-approximated Lie Algebra for Shooting~(FLASH), a fast variant of LDDMM with geodesic shooting~\cite{zhang2019fast}. Given a set of images $I_1,\cdots,I_N$ with $N$ being the number of images, the problem of atlas building is to find a template image $I$ and transformations $\phi_1,\cdots, \phi_N$ that minimize the energy function
\begin{equation}
\label{eq:lddmm}
E(I, \phi_n) = \sum_{n=1}^{N} \text{Dist} (I \circ \phi_n, I_n) + \text{Reg}(\alpha, \phi_n). 
\end{equation} 
The $\text{Dist}(\cdot, \cdot)$ is a distance function that measures the dissimilarity between images, i.e., sum-of-squared differences~\cite{beg2005computing}, normalized cross correlation~\cite{avants2008symmetric}, and mutual information~\cite{wells1996multi}. The $\Reg(\cdot)$ is a weighted regularization with parameter $\alpha$ that guarantees the diffeomorphic properties of transformation fields. 

\paragraph*{\bf Regularization In Tangent Space of Diffeomorphisms.}
Given an open and bounded $d$-dimensional domain $\Omega \subset \mathbb{R}^d$, we use $\Diff(\Omega)$ to denote a space of diffeomorphisms and its tangent space $V=T \Diff(\Omega)$. The regularization of LDDMM is defined as an integral of the Sobolev norm of the time-dependent velocity field $v(t) \in V  (t \in [0, 1])$ in the tangent space, i.e., 
\begin{equation}
\label{eq:lddmm}
\text{Reg}(\alpha, \phi_n) = \int \langle \L(\alpha) v_n(t), \L(\alpha) v_n(t) \rangle \, dt, \, \text{with} \, \, \frac{d\phi_n(t)}{dt} = - D\phi_n(t)\cdot v_n(t).
\end{equation}
Here $\L$ is a symmetric, positive-definite differential operator, with parameter $\alpha$ controling the smoothness of transformation fields. In this paper, we use the Laplacian operator $\L=(- \alpha \Delta + \text{Id})^3$, where $\text{Id}$ is an identity matrix. The operator $D$ is a Jacobian matrix and $\cdot$ denotes an element-wise matrix multiplication.

According to the geodesic shooting algorithm~\cite{vialard2012}, the minimum of LDDMM is uniquely determined by solving a Euler-Poincar\'{e} differential equation (EPDiff)~\cite{arnold1966,miller2006geodesic} with initial conditions. This inspires a recent model FLASH to reparameterize the regularization of Eq.~\eqref{eq:lddmm} in a low-dimensional bandlimited space of initial velocity fields, which dramatically reduces the computational complexity of transformation models with little to no loss of accuracy~\cite{zhang2019fast}.  

\paragraph*{\bf Fourier Computation of Diffeomorphisms.} Let $\widetilde{\Diff}(\Omega)$ and $\tilde{V}$ denote the space of Fourier representations of diffeomorphisms
and velocity fields respectively. Given time-dependent velocity field $\tilde{v}(t) \in \tilde{V}$, the diffeomorphism $\tilde{\phi}(t) \in \widetilde{\Diff}(\Omega)$ in the finite-dimensional Fourier domain can be computed as
\begin{align}\label{eq:leftinvariantfft}
\tilde{\phi}(t) = \tilde{\text{Id}} + \tilde{u}(t), \quad \frac{d \tilde{u}(t)}{dt} &= -\tilde{v}(t) - \tilde{\mathcal{D}} \tilde{u}(t) \ast \tilde{v}(t),
\end{align}
where $\tilde{\text{Id}}$ is the frequency of an identity element, $\tilde{\mathcal{D}}\tilde{u}(t)$ is a tensor product $\tilde{\mathcal{D}} \otimes \tilde{u}(t)$, representing the Fourier frequencies of a Jacobian matrix $\tilde{\mathcal{D}}$ with central difference approximation, and $\ast$ is a circular convolution~\footnote{To prevent the domain from growing infinity, we truncate the output of the convolution in each dimension to a suitable finite set.}.

The Fourier representation of the geodesic shooting equation (EPDiff) is
\begin{align}\label{eq:epdiffleft}
    \frac{\partial \tilde{v}(t)}{\partial t} =-\tilde{\K}\left[(\tilde{\mathcal{D}} \tilde{v}(t))^T \star \tilde{\mathcal{\L}}\tilde{v}(t) + \tilde{\nabla} \cdot (\tilde{\mathcal{\L}}\tilde{v}(t) \otimes \tilde{v}(t)) \right],
\end{align}
where $\star$ is the truncated matrix-vector field auto-correlation. The operator $\tilde{\nabla} \cdot$ is the discrete divergence of a vector field. Here $\tilde{\K}$ is an inverse operator of $\tilde{\L}$, which is the Fourier transform of a Laplacian operator in this paper.

The regularization in Eq.~\eqref{eq:lddmm} can be equivalently formulated as
\begin{align*}
  \text{Reg}(\alpha, \phi_n) = \langle \tilde{{\L}}(\alpha) \tilde{v}_n(0), \tilde{{\L}}(\alpha) \tilde{v}_n(0) \rangle, \quad \text{s.t.} \, \text{Eq.}~\eqref{eq:leftinvariantfft} \& \text{Eq.}~\eqref{eq:epdiffleft}.
\end{align*}
We will drop off the time index in remaining sections for notational simplicity, e.g., defining $\tilde{v}_n \triangleq \tilde{v}_n(0)$.   


\section{Our Model: Bayesian Atlas Building with Hierarchical Priors}
This section presents a hierarchical Bayesian model for atlas building that allows subject-specific regularization with no manual effort of parameter-tuning. We introduce a hierarchical prior distribution on the initial velocity fields with adaptive smoothing parameters followed by a likelihood distribution on images.  

\paragraph*{\bf Likelihood.} Assuming an independent and identically distributed (i.i.d.) Gaussian noise on image intensities, we formulate the likelihood of each observed image $I_n$ as  
\begin{align}
\label{eq:likelihood}
 p(I_n \, | \, I, \tilde{v}_n, \sigma^2) = \frac{1}{(\sqrt{2 \pi} \sigma ^2 )^{M}} \exp{ \left(-\frac{1}{2\sigma ^2} \lVert I \circ \phi_n - I_n \rVert_2^2 \right)}. 
\end{align} 
Here $\sigma^2$ denotes a noise variance, $M$ is the number of image voxels, and $\phi_n$ is an inverse Fourier transform of $\tilde{\phi}_n$ at time point $t=1$. It is worth mentioning that other noise models such as spatially varying noises~\cite{simpson2012probabilistic} can also be applied.  

\paragraph*{\bf Prior.} To ensure the smoothness of transformation fields, we define a prior on each initial velocity field $\tilde{v}_n$ as a complex multivariate Gaussian distribution
\begin{align}
\label{eq:prior}
p({\tilde{v}}_n \, | \, \alpha_n ) = \frac{1}{(2 \pi)^{\frac{M}{2}} | {\tilde{\L}_n}^{-1}(\alpha_n) |} \exp\left ({-\frac{1}{2} \langle {\tilde{\L}_n(\alpha_n)} {\tilde{v}}_n, {\tilde{\L}(\alpha_n)} {\tilde{v}}_n \rangle} \right),
\end{align}
where $|\cdot|$ is matrix determinant. The Fourier coefficients of a discrete Laplacian operator is $\tilde{{\L}}_n(\xi_1 , \ldots, \xi_d) = \left(-2 \alpha_n \sum_{j = 1}^d \left(\cos (2\pi \xi_j) - 1 \right) + 1\right)^3$, with $(\xi_1 , \ldots, \xi_d)$ being a d-dimensional frequency vector. 

\paragraph*{\bf Hyperprior.} We treat the subject-specific regularization parameter $\alpha_n$ of the prior distribution~\eqref{eq:prior} as a random variable generated from Gamma distribution, which is a commonly used prior to model positive real numbers~\cite{simpson2015probabilistic}. Other prior such as inverse Wishart distribution~\cite{gori2017bayesian} can also be applied. The hyperprior of our model is formulated as
\begin{align}
\label{eq:hyperprior}
p(\alpha_n \, | \, k, \beta) = \frac{\alpha_n^{k-1} \exp ^{(-\alpha_n / \beta)}}{\Gamma (k) \beta^{k}},
\end{align}
with $k$ and $\beta$ being positive numbers for shape and scale parameters respectively. The Gamma function $\Gamma(k)=(k-1)!$ for all positive integers of $k$.
We finally arrive at the log posterior of the diffeomorphic transformation and regularization parameters as
\begin{align}
E(\tilde{v}_n, \alpha_n, I, \sigma, k, \beta) & \triangleq \ln \prod_{n=1}^N  p(I_n \, | \, I, \tilde{v}_n, \sigma^2) \cdot p({\tilde{v}}_n \, | \, \alpha_n ) \cdot p(\alpha_n \, | \, k, \beta) \nonumber \\  
&= \sum_{n=1}^{N} \frac{1}{2}\ln\lvert \L_n \rvert - M \ln \sigma-\frac{ \| I \circ \phi_n -I_n \|_2^2}{2\sigma^2} - \frac{1}{2}(\tilde{\L} \tilde{v}_n, \tilde{\L} \tilde{v}_n)  \nonumber \\  
& \quad (k-1) \ln \alpha_n  -\frac{\alpha_n}{\beta} - k \ln \beta - \ln \Gamma(k)   +\text{const.} 
\label{eq:posterior}
\end{align}

\subsection{Model Inference}

We develop an MCEM algorithm to infer the model parameter $\Theta$, which includes the image atlas $I$, the noise variance of image intensities $\sigma^2$, the initial velocities of diffeomorphic transformations $\tilde{v}_n$, and the hyperparameters $k$ and $\beta$. We treat the regularization parameter $\alpha_n$ as latent random variables and integrate them out from the log posterior in Eq.~\eqref{eq:posterior}.  
Computations of two main steps (expectation and maximization) are illustrated below. 

\paragraph*{\bf Expectation: HMC.} Since the E-step does not yield a closed-form solution, we employ a powerful Hamiltonian Monte Carlo (HMC) sampling method~\cite{duane1987hybrid} to approximate the expectation function $Q$ with respect to the latent variables $\alpha_n$. For each $\alpha_n$, we draw a number of $S$ samples from the log posterior~\eqref{eq:posterior} by using HMC from the current estimated parameters $\hat{\Theta}$. The Monte Carlo approximation of the expectation $Q$ is
\begin{align}
Q(\Theta | \hat{\Theta}) \approx \frac{1}{S} \sum_{n=1}^{N} \sum_{j=1}^{S} \ln p(\alpha_{nj} \, | \, I_n; \hat{\Theta}).
 \label{eq:expectation}
\end{align}

To produce samples of $\alpha_n$, we first define the potential energy of the Hamiltonian system $H(\alpha_n,\gamma) = U(\alpha_n)+W(\gamma)$ as $U(\alpha_n) = -\ln p(\alpha_n | I_n; \Theta)$. The kinetic energy $W(\gamma)$ is a typical normal distribution on an auxiliary variable $\gamma$. This gives us Hamilton's equations to integrate
\begin{align}\label{eq:hmc}
\frac{\alpha_n}{dt} = \frac{\partial H}{\partial \gamma} = \gamma, \quad
\frac{d \gamma}{dt} &=-\frac{\partial H}{\partial \alpha_n}=- \nabla_{\alpha_n}U. 
\end{align}
Since $\alpha_n$ is a Euclidean variable, we use a standard ``leap-frog'' numerical integration scheme, which approximately conserves the Hamiltonian and results in high acceptance rates. The gradient of $U$ with respect to $\alpha_{n}$ is
\begin{align}
    \nabla_{\alpha_{n}}U = \frac{3}{2S}\sum_{j=1}^S[ \sum_{i=1}^d \frac{\tilde{\A}_{i}}{\alpha_{nj} \tilde{A}_i + 1} - \langle 2(\alpha_{nj} \tilde{\A} +1)^{5}\tilde{\A} \tilde{v}_{nj}, \tilde{v}_{nj} \rangle],
    \label{eq:alphagrad}
\end{align}
where $\tilde{\A}= -2 \sum_{i = 1}^d \left(\cos (2\pi \xi_i) - 1 \right)$. Here $\tilde{\A}$ denotes a discrete Fourier Laplacian operator with a $d$-dimensional frequency vector.

Starting from the current point $\alpha_n$ and initial random auxiliary variable $\gamma$, the Hamiltonian system is integrated forward in time by Eq.~\eqref{eq:hmc} to produce a candidate point $(\hat{\alpha}_n, \hat{\gamma})$. The candidate point $\hat{\alpha}_n$ is accepted as a new point in the sample with probability $p(accept) = \min(1, -U(\hat{\alpha}_n)-W(\hat{\gamma}) + U(\alpha_n)+W(\gamma))$.

\paragraph*{\bf Maximization: Gradient Ascent.} We derive the maximization step to update the parameters $\Theta=\{I, \tilde{v}_n, \sigma^2, k, \beta \}$ by maximizing the HMC approximation of the expectation $Q$ in Eq.~\eqref{eq:expectation}. 

For updating the atlas image $I$, we set the derivative of the $Q$ function with respect to $I$ to zero. The solution for $I$ gives a closed-form update
\begin{align}
     I = \frac{\sum_{j=1}^S \sum_{n=1}^N (I_n \circ \phi^{-1}_{nj}) \cdot \lvert D \phi^{-1}_{nj} \rvert }{\sum_{j=1}^S \sum_{n=1}^N \lvert D \phi^{-1}_{nj} \rvert}.
     \label{eq:atlasupdate}
\end{align}

Similarly, we obtain the closed-form solution for the noise variance $\sigma^2$ after setting the gradient of $Q$ w.r.t. $\sigma^2$ to zero 
\begin{align}
    \sigma^2 = \frac{1}{MNS} \sum_{n=1}^N\sum_{j=1}^S \lVert I \circ \phi_{nj}-I_n \rVert_2^2. 
    \label{eq:sigmaupdate}
\end{align}

The closed-form solutions for hyperparameters $k$ and $\beta$ are
\begin{align}
k = \psi^{-1}( \frac{1}{NS}\sum_{i=1}^{N} \sum_{j = 1}^{S} \ln \alpha_{nj}-\ln \beta ), \quad \beta = \frac{1}{NS k}\sum_{n=1}^N \sum_{j = 1}^S \alpha_{nj}.
\label{eq:betakupdate}
\end{align}
Here $\psi$ is a digamma function, which is the logarithmic derivative of the gamma function $\Gamma (\cdot)$. The inverse of digamma function $\psi^{-1}$ is computed by using a fixed-point iteration algorithm~\cite{minka2000estimating}.

As there is no closed-form update for initial velocities, we employ a gradient ascent algorithm to estimate $\tilde{v}_{nj}$. The gradient $\nabla_{\tilde{v}_{nj}} Q$ is computed by a forward-backward sweep approach. Details are introduced in the FLASH algorithm~\cite{zhang2019fast}. 

\section{Experimental Evaluation}
We compare the proposed model with LDDMM atlas building algorithm that employs single-penalty regularization with manually tuned parameters on 3D brain images~\cite{zhang2019fast}. In HMC sampling, we draw $300$ samples for each subject, with initialized value of $\alpha = 10$, $k = 9.0$, $\sigma=0.05$, and $\beta = 0.1$. An averaged image of all image intensities is used for atlas initialization.

\noindent{\bf Data.} We include $100$ 3D brain MRI scans with segmentation maps from a public released resource Open Access Series of Imaging Studies (OASIS) for Alzheimer's disease~\cite{fotenos2005normative}. The dataset covers both healthy and diseased subjects, aged from $55$ to $90$. The MRI scans are resampled to $128^3$ with the voxel size of $1.25 mm^3$. All MRIs are carefully prepossessed by skull-stripping, intensity normalization, bias field correction, and co-registration with affine transformation. 

\noindent{\bf Experiments.} We estimate the atlas of all deformed images by using our method and compare its performance with LDDMM atlas building~\cite{zhang2019fast}. Final results of atlases estimated from both our model and the baseline algorithm are reported. We also compare the time and memory consumption of proposed model with the baseline that performs HMC sampling in a full spatial domain~\cite{zhang2013bayesian}. To measure the sharpness of estimated atlas $I$, we adopt a metric of normalized standard deviation computed from randomly selected 3000 image patches~\cite{legouhy2019online}. Given $N(i)$,
a patch around a voxel $i$ of an atlas $I$, the local measure of the sharpness at voxel $i$ is defined as $\text{sharpness}(I(i)) = \text{sd}_{N(i)}(I)/ \text{avg}_{N(i)}(I)$, where sd and avg denote the standard deviation and the mean of $N_i$. 

To further evaluate the quality of estimated transformations, we perform atlas-based segmentation after obtaining transformations from our model. For a fair comparison, we fix the atlas for both methods and examine the registration accuracy by computing the dice similarity coefficient (DSC)~\cite{dice1945measures} between the propagated segmentation and the manual segmentation on six anatomical brain structures, including cerebellum white matter, thalamus, brain stem, lateral ventricle, putamen, caudate. The significance tests on both dice and sharpness between our method and the baseline are performed. 

\begin{figure}[!ht]
\begin{center}
 \includegraphics[width=0.96\textwidth] {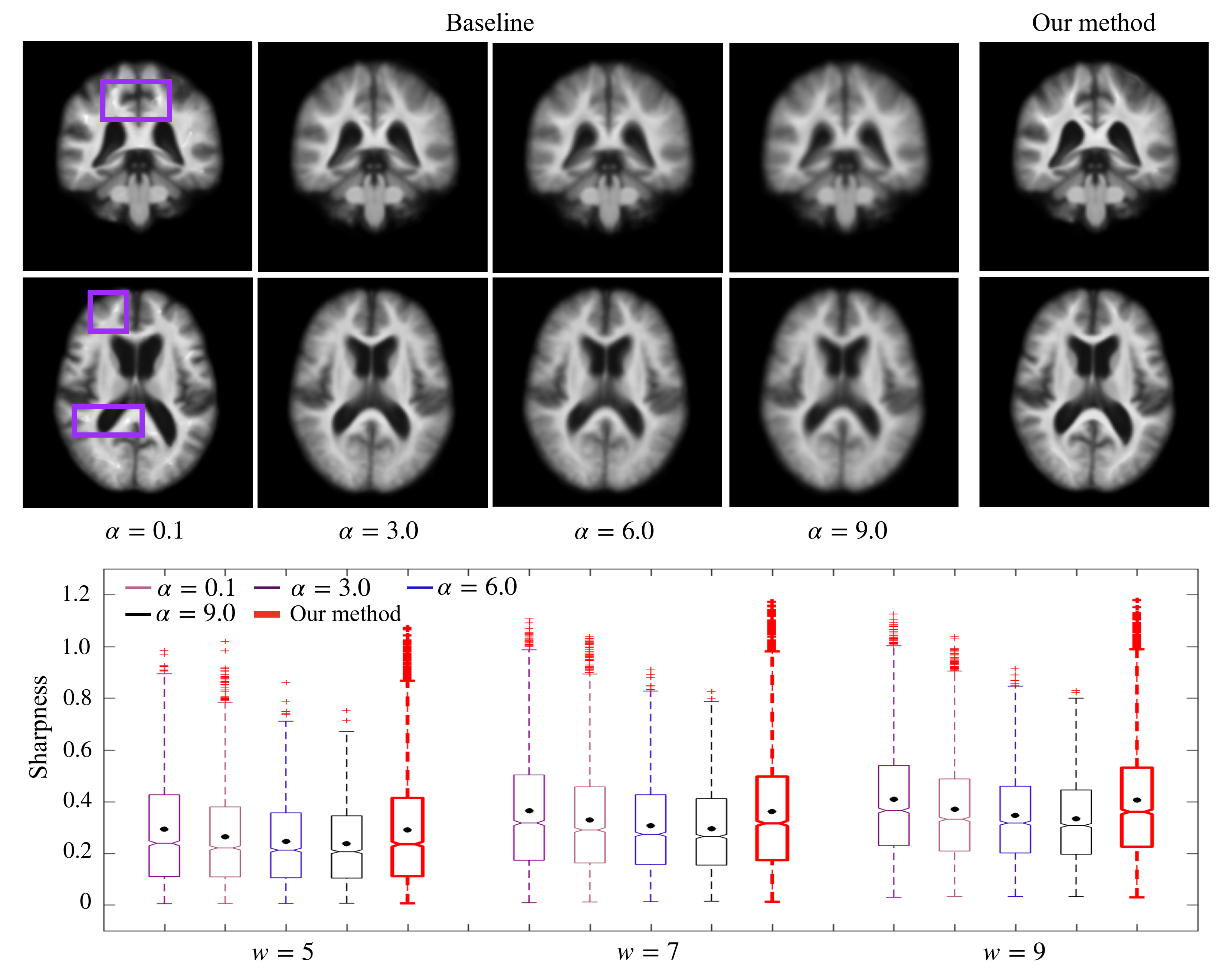}
 \caption{Top: atlases estimated by baseline with different $\alpha$ and our model (artifacts introduced by small regularization are outlined in purple boxes). Bottom: sharpness measurement of atlas for all methods with different patch size $w$. The mean of the sharpness metric of \textbf {our method} vs. the best performance of baseline without artifacts ($\alpha =3$) is \textbf{0.290}/0.264,  \textbf{0.362}/0.323, \textbf{0.405}/0.360. }
\label{fig:3datlas}
\end{center}             
\end{figure}

\noindent{\bf Results.} Fig.~\ref{fig:3datlas} visualizes a comparison of 3D atlas on real brain MRI scans. The top panel shows that our model substantially improves the quality of atlas with sharper and better details than the baseline with different values of manually set regularization parameters, e.g., $\alpha=0.1, 3.0, 6.0, 9.0$. Despite the observation of a smaller value of $\alpha=0.1$ produces sharper atlas, it breaks the smoothness constraints on the transformation fields hence introducing artifacts on anatomical structures (outlined in purple boxes). The mean and standard deviation of our estimated hyperprior parameters $k$ and $\beta$ in Eq.~\eqref{eq:hyperprior} over $30$ pairwise image registrations are $47.40/7.22$, and $0.036/0.005$. The bottom panel quantitatively reports the sharpness metric of all methods. It indicates that our algorithm outperforms the baseline by offering a higher sharpness score while preserving the topological structure of brain anatomy. 

\begin{figure}[!h]
\begin{center}
 \includegraphics[width=.96\textwidth] {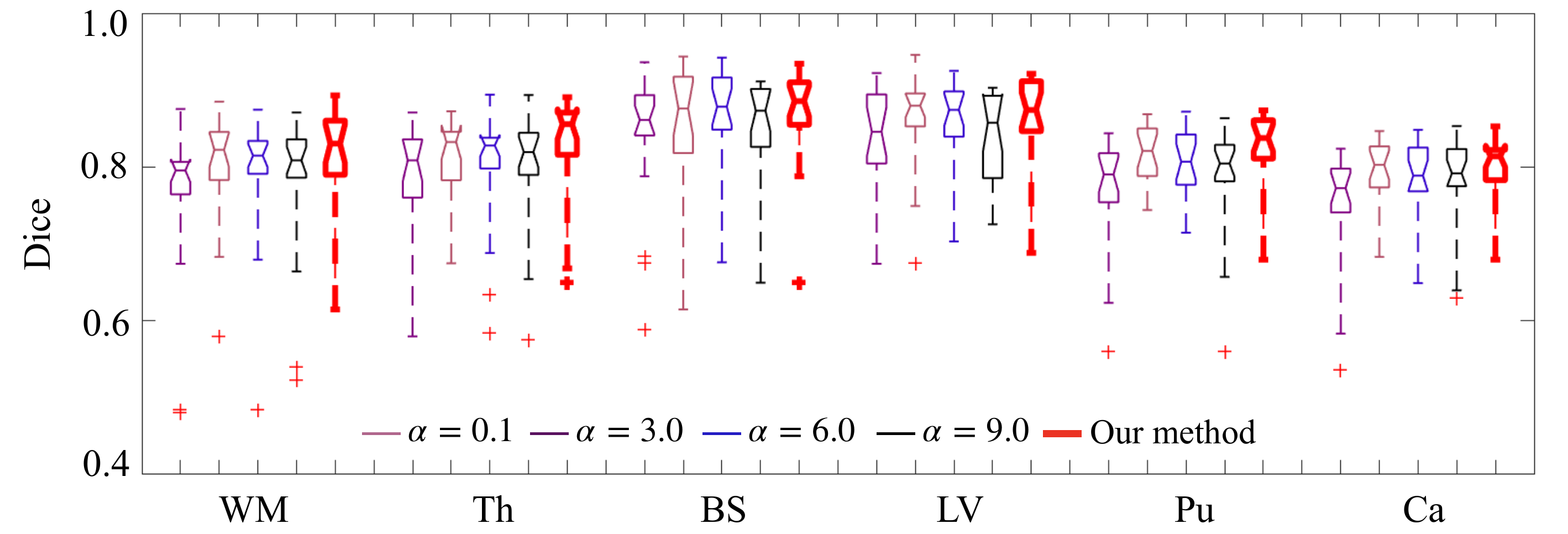}
     \caption{A comparison of dice evaluation for fixed-atlas-based segmentation on six brain structures (cerebellum white matter (WM), thalamus (Th), brain stem (BS), lateral ventricle (LV), putamen(Pu), caudate (Ca)).}
\label{fig:dice}
\end{center}             
\end{figure}

Fig.~\ref{fig:dice} reports results of fixed-atlas-based segmentation by performing the baseline with various regularization parameters and our algorithm. It shows the dice comparison on six anatomical brain structures of all image pairs. Our algorithm produces better dice coefficients without the need of parameter tuning. 

The runtime of our atlas building on $100$ 3D brain MR images are $4.4$ hours with $0.89$GB memory consumption. The p-values of significance differences test on both dice ($p=0.002$) and sharpness ($p=0.0034$) reject the null hypothesis that there's no differences between our model estimation and baseline algorithms.
\section{Conclusion}
This paper presents a novel hierarchical Bayesian model for unbiased diffeomorphic atlas building with subject-specific regularization. We design a new parameter choice rule that allows adaptive regularization to control the smoothness of image transformations. We introduce a hierarchical prior that provides prior information of regularization parameters at multiple levels. The developed MCEM inference algorithm eliminates the need of manual parameter tuning, which can be tedious and infeasible in multi-parameter settings. Experimental results show that our proposed algorithm yields a better registration model as well as an improved quality of atlas. While our algorithm is presented in the setting of LDDMM, the theoretical development is generic to other deformation models, e.g., stationary velocity fields~\cite{arsigny2006log}. In addition, this model can be easily extended to multi-atlas building where a much higher degree of variations exist in the population studies. Our future work will focus on conducting subsequent statistical shape analysis in the resulting atlas space.

\appendix
\bibliographystyle{splncs04}
\bibliography{miccaipaper}

\begin{thebibliography}{10}
\providecommand{\url}[1]{\texttt{#1}}
\providecommand{\urlprefix}{URL }
\providecommand{\doi}[1]{https://doi.org/#1}

\bibitem{allassonniere2007towards}
Allassonni{\`e}re, S., Amit, Y., Trouv{\'e}, A.: Towards a coherent statistical
  framework for dense deformable template estimation. Journal of the Royal
  Statistical Society: Series B (Statistical Methodology)  \textbf{69}(1),
  3--29 (2007)

\bibitem{allassonniere2008stochastic}
Allassonni{\`e}re, S., Kuhn, E.: Stochastic algorithm for parameter estimation
  for dense deformable template mixture model. arXiv preprint arXiv:0802.1521
  (2008)

\bibitem{arnold1966}
Arnol'd, V.I.: Sur la g\'{e}om\'{e}trie diff\'{e}rentielle des groupes de {L}ie
  de dimension infinie et ses applications \`{a} l'hydrodynamique des fluides
  parfaits. Ann. Inst. Fourier  \textbf{16},  319--361 (1966)

\bibitem{arsigny2006log}
Arsigny, V., Commowick, O., Pennec, X., Ayache, N.: A log-euclidean framework
  for statistics on diffeomorphisms. In: International Conference on Medical
  Image Computing and Computer-Assisted Intervention. pp. 924--931. Springer
  (2006)

\bibitem{ashburner2011diffeomorphic}
Ashburner, J., Friston, K.J.: Diffeomorphic registration using geodesic
  shooting and gauss--newton optimisation. NeuroImage  \textbf{55}(3),
  954--967 (2011)

\bibitem{avants2008symmetric}
Avants, B.B., Epstein, C.L., Grossman, M., Gee, J.C.: Symmetric diffeomorphic
  image registration with cross-correlation: evaluating automated labeling of
  elderly and neurodegenerative brain. Medical image analysis  \textbf{12}(1),
  26--41 (2008)

\bibitem{beg2005computing}
Beg, M.F., Miller, M.I., Trouv{\'e}, A., Younes, L.: Computing large
  deformation metric mappings via geodesic flows of diffeomorphisms.
  International journal of computer vision  \textbf{61}(2),  139--157 (2005)

\bibitem{dice1945measures}
Dice, L.R.: Measures of the amount of ecologic association between species.
  Ecology  \textbf{26}(3),  297--302 (1945)

\bibitem{duane1987hybrid}
Duane, S., Kennedy, A.D., Pendleton, B.J., Roweth, D.: Hybrid monte carlo.
  Physics letters B  \textbf{195}(2),  216--222 (1987)

\bibitem{fotenos2005normative}
Fotenos, A.F., Snyder, A., Girton, L., Morris, J., Buckner, R.: Normative
  estimates of cross-sectional and longitudinal brain volume decline in aging
  and ad. Neurology  \textbf{64}(6),  1032--1039 (2005)

\bibitem{gori2017bayesian}
Gori, P., Colliot, O., Marrakchi-Kacem, L., Worbe, Y., Poupon, C., Hartmann,
  A., Ayache, N., Durrleman, S.: A bayesian framework for joint morphometry of
  surface and curve meshes in multi-object complexes. Medical image analysis
  \textbf{35},  458--474 (2017)

\bibitem{hong2017fast}
Hong, Y., Golland, P., Zhang, M.: Fast geodesic regression for population-based
  image analysis. In: International Conference on Medical Image Computing and
  Computer-Assisted Intervention. pp. 317--325. Springer (2017)

\bibitem{iglesias2012incorporating}
Iglesias, J.E., Sabuncu, M.R., Van~Leemput, K.: Incorporating parameter
  uncertainty in bayesian segmentation models: Application to hippocampal
  subfield volumetry. In: International Conference on Medical Image Computing
  and Computer-Assisted Intervention. pp. 50--57. Springer (2012)

\bibitem{joshi2004unbiased}
Joshi, S., Davis, B., Jomier, M., Gerig, G.: Unbiased diffeomorphic atlas
  construction for computational anatomy. NeuroImage  \textbf{23},  S151--S160
  (2004)

\bibitem{legouhy2019online}
Legouhy, A., Commowick, O., Rousseau, F., Barillot, C.: Online atlasing using
  an iterative centroid. In: International Conference on Medical Image
  Computing and Computer-Assisted Intervention. pp. 366--374. Springer (2019)

\bibitem{liao2019temporal}
Liao, R., Turk, E.A., Zhang, M., Luo, J., Adalsteinsson, E., Grant, P.E.,
  Golland, P.: Temporal registration in application to in-utero mri time
  series. arXiv preprint arXiv:1903.02959  (2019)

\bibitem{lorenzo2002atlas}
Lorenzo-Vald{\'e}s, M., Sanchez-Ortiz, G.I., Mohiaddin, R., Rueckert, D.:
  Atlas-based segmentation and tracking of 3d cardiac mr images using non-rigid
  registration. In: International conference on medical image computing and
  computer-assisted intervention. pp. 642--650. Springer (2002)

\bibitem{ma2008bayesian}
Ma, J., Miller, M.I., Trouv{\'e}, A., Younes, L.: Bayesian template estimation
  in computational anatomy. NeuroImage  \textbf{42}(1),  252--261 (2008)

\bibitem{miller2006geodesic}
Miller, M.I., Trouv{\'e}, A., Younes, L.: Geodesic shooting for computational
  anatomy. Journal of mathematical imaging and vision  \textbf{24}(2),
  209--228 (2006)

\bibitem{minka2000estimating}
Minka, T.: Estimating a dirichlet distribution (2000)

\bibitem{pohl2006bayesian}
Pohl, K.M., Fisher, J., Grimson, W.E.L., Kikinis, R., Wells, W.M.: A bayesian
  model for joint segmentation and registration. NeuroImage  \textbf{31}(1),
  228--239 (2006)

\bibitem{rohlfing2004evaluation}
Rohlfing, T., Brandt, R., Menzel, R., Maurer~Jr, C.R.: Evaluation of atlas
  selection strategies for atlas-based image segmentation with application to
  confocal microscopy images of bee brains. NeuroImage  \textbf{21}(4),
  1428--1442 (2004)

\bibitem{simpson2015probabilistic}
Simpson, I.J., Cardoso, M.J., Modat, M., Cash, D.M., Woolrich, M.W., Andersson,
  J.L., Schnabel, J.A., Ourselin, S., Initiative, A.D.N., et~al.: Probabilistic
  non-linear registration with spatially adaptive regularisation. Medical image
  analysis  \textbf{26}(1),  203--216 (2015)

\bibitem{simpson2012probabilistic}
Simpson, I.J., Woolrich, M.W., Andersson, J.L., Groves, A.R., Schnabel, J.A.: A
  probabilistic non-rigid registration framework using local noise estimates.
  In: 2012 9th IEEE International Symposium on Biomedical Imaging (ISBI). pp.
  688--691. IEEE (2012)

\bibitem{vialard2012}
Vialard, F.X., Risser, L., Rueckert, D., Cotter, C.J.: Diffeomorphic 3d image
  registration via geodesic shooting using an efficient adjoint calculation.
  International Journal of Computer Vision  \textbf{97}(2),  229--241 (2012)

\bibitem{vialard2011diffeomorphic}
Vialard, F.X., Risser, L., Holm, D.D., Rueckert, D.: Diffeomorphic atlas
  estimation using karcher mean and geodesic shooting on volumetric images. In:
  MIUA. pp. 55--60 (2011)

\bibitem{wang2019data}
Wang, J., Xing, W., Kirby, R.M., Zhang, M.: Data-driven model order reduction
  for diffeomorphic image registration. In: International conference on
  information processing in medical imaging. pp. 694--705. Springer (2019)

\bibitem{wells1996multi}
Wells~III, W.M., Viola, P., Atsumi, H., Nakajima, S., Kikinis, R.: Multi-modal
  volume registration by maximization of mutual information. Medical image
  analysis  \textbf{1}(1),  35--51 (1996)

\bibitem{yeo2008effects}
Yeo, B.T., Sabuncu, M.R., Desikan, R., Fischl, B., Golland, P.: Effects of
  registration regularization and atlas sharpness on segmentation accuracy.
  Medical image analysis  \textbf{12}(5),  603--615 (2008)

\bibitem{zhang2019fast}
Zhang, M., Fletcher, P.T.: Fast diffeomorphic image registration via
  fourier-approximated lie algebras. International Journal of Computer Vision
  \textbf{127}(1),  61--73 (2019)

\bibitem{zhang2013bayesian}
Zhang, M., Singh, N., Fletcher, P.T.: Bayesian estimation of regularization and
  atlas building in diffeomorphic image registration. In: Information
  Processing in Medical Imaging. pp. 37--48. Springer (2013)

\end{thebibliography}
\end{document}